\documentclass{ecai}  % use option [doubleblind] for double blind submission and hiding the authors section

\usepackage{graphicx}
\usepackage{latexsym}

\ecaisubmission      % inserts page numbers. Use only for submission of paper.
\usepackage{mathtools}
\usepackage{amsfonts}       % blackboard math symbols
\usepackage{amsmath,bm}
\usepackage{amssymb}
\usepackage{algorithm}
\usepackage{dblfloatfix}
\usepackage{algpseudocode}
\newcommand{\vect}[1]{\boldsymbol{\mathbf{#1}}}
\newcommand{\Ex}{\mathbb{E}}

\newcommand{\Wspace}{\mathcal{W}}

\begin{document}

\begin{frontmatter}

\title{Hyperparameter Optimization for\\ Multi-Objective Reinforcement Learning}

\author[A]{\fnms{Florian}~\snm{Felten}\orcid{0000-0002-2874-3645}\thanks{Corresponding Author. Email: florian.felten@uni.lu.}}
\author[B]{\fnms{Daniel}~\snm{Gareev}}
\author[A,C]{\fnms{El-Ghazali}~\snm{Talbi}\orcid{0000-0003-4549-1010}}
\author[A,B]{\fnms{Grégoire}~\snm{Danoy}\orcid{0000-0001-9419-4210}} % use of \orcid{} is optional

\address[A]{SnT, University of Luxembourg}
\address[B]{FSTM/DCS, University of Luxembourg}
\address[C]{CNRS/CRIStAL, University of Lille}

\begin{abstract}
Reinforcement learning (RL) has emerged as a powerful approach for tackling complex problems. The recent introduction of multi-objective reinforcement learning (MORL) has further expanded the scope of RL by enabling agents to make trade-offs among multiple objectives. This advancement not only has broadened the range of problems that can be tackled but also created numerous opportunities for exploration and advancement. Yet, the effectiveness of RL agents heavily relies on appropriately setting their hyperparameters. In practice, this task often proves to be challenging, leading to unsuccessful deployments of these techniques in various instances. Hence, prior research has explored hyperparameter optimization in RL to address this concern. 

This paper presents an initial investigation into the challenge of hyperparameter optimization specifically for MORL. We formalize the problem, highlight its distinctive challenges, and propose a systematic methodology to address it. The proposed methodology is applied to a well-known environment using a state-of-the-art MORL algorithm, and preliminary results are reported. Our findings indicate that the proposed methodology can effectively provide hyperparameter configurations that significantly enhance the performance of MORL agents. Furthermore, this study identifies various future research opportunities to further advance the field of hyperparameter optimization for MORL.
\end{abstract}

\end{frontmatter}
\vspace{-2mm}
\section{Introduction}

RL has become a popular approach to solving challenging problems in recent years. This is due in part to its success in a range of domains, such as playing Atari games~\cite{mnih_human-level_2015} and Go~\cite{silver_mastering_2016}. Within RL, the domain of multi-objective RL (MORL) has recently emerged, where agents are tasked with making compromises between multiple rewards.

The field of MORL has seen the development of several algorithms and important theoretical findings. These can be found in recent surveys such as~\cite{roijers_survey_2013,hayes_practical_2022}. Additionally, software tools have been released to facilitate reproducibility, adoption, and further advancement of MORL methods~\cite{alegre_mo-gym_2022,felten_toolkit_2023}. %For instance, MO-Gymnasium offers a collection of MORL environments under a standardized API. MORL-Baselines~\cite{felten_morl-baselines_2022}, compatible with MO-Gymnasium, provides a range of MORL algorithms and development tools.

Even though open-source implementations and tools are readily accessible, promoting the use of these techniques even among non domain experts, it is widely recognized that the effectiveness of RL (and MORL) agents is highly dependent on the values of their hyperparameters~\cite{henderson_deep_2018,andrychowicz_what_2021,patterson_empirical_2023}. This implies that in a new environment, an existing algorithm may exhibit poor performance when run with its default hyperparameters. Furthermore, baseline comparisons frequently involve contrasting new algorithms with state-of-the-art techniques. In this regard, \cite{patterson_empirical_2023}~pinpoint the problem of unfairness when untuned baseline algorithms are pitted against finely tuned new algorithms. 

While manually tuning these hyperparameters may seem appealing, the large number of parameters and the time required to obtain training results often make manual optimization impractical~\cite{parker-holder_automated_2022}. 
Consequently, having an automated approach to tuning hyperparameters for MORL, given a predefined level of effort, not only facilitates fairer comparisons within the field but also streamlines the implementation of these techniques for real-world applications.
%Ultimately, employing sensitivity analysis tools to assess the impact of hyperparameters could enable a comprehensive examination and comprehension of algorithm behavior~\cite{parker-holder_automated_2022,bischl_hyperparameter_2023}.
Yet, to the best of our knowledge, although recent endeavors have focused on automating such decisions in the context of RL~\cite{parker-holder_automated_2022,eimer_hyperparameters_2023}, no similar effort has been applied to MORL. Moreover, existing solutions for single-objective RL do not trivially apply to MORL.

The objective of this work is to address this gap by formally recognizing the challenge of hyperparameter optimization for MORL and introducing a systematic procedure to undertake such an endeavor. Section 2 introduces the necessary background knowledge and notation related to MORL and hyperparameter optimization (HPO). Section 3 formally defines the problem of hyperparameter optimization for multi-objective RL and proposes a methodology to solve it. Section 4 showcases preliminary results of the proposed methodology applied to a state-of-the-art method. Lastly, Section 5 provides a conclusion to our research and discusses potential leads for future work.
\vspace{-2mm}
\section{Background}

This section introduces notations and background information for multi-objective reinforcement learning and hyperparameter optimization.
\vspace{-3mm}
\subsection{Multi-objective reinforcement learning}
\label{sec:MORL}

MORL utilizes the multi-objective Markov decision process (MOMDP) framework to capture an agent's interaction with its environment~\cite{roijers_survey_2013,hayes_practical_2022}. This framework differs from Markov decision processes (MDPs)~\cite{sutton_reinforcement_2018} in that it expands the reward function to generate vectors instead of scalar values.
Formally, a MOMDP is defined as a tuple $(\mathcal{S},\mathcal{A},p,\vect{r},\mu,\gamma)$, where $\mathcal{S}$ represents a state space,  $\mathcal{A}$ an action space, $p: \mathcal{S} \times \mathcal{A}\times  \mathcal{S} \mapsto [0, 1]$ is the distribution over next states given state $s$ and action $a$, $\vect{r} :\mathcal{S} \times \mathcal{A} \times \mathcal{S} \mapsto \mathbb{R}^m$ is a multi-objective reward function containing $m$ objectives, $\mu$ is an initial state distribution, and $\gamma \in [0,1)$ is a discounting factor. A \textit{policy} $\pi : \mathcal{S} \mapsto \mathcal{A}$ is a function mapping states to actions, which is typically what the agent aims at learning.
Let $S_t$ and $A_t$ denote the variables corresponding to state and action at time step $t$.
The vector value function of a policy $\pi$ in state $s$ is defined as 
\begin{equation*}
    \vect{v}^\pi(s) \equiv \Ex_\pi \left[\sum_{t=0}^{\infty} \gamma^t \vect{r}(S_{t},A_{t},S_{t+1}) \ | \ S_t = s \right],
\end{equation*}
where $\Ex_\pi[\cdot]$ denotes expectation with respect to trajectories induced by the policy $\pi$.
We denote $\vect{v}^\pi \equiv \Ex_{s_0\sim\mu}[\vect{v}^\pi(s_0)]$ as the \emph{value vector} of policy $\pi$. 
%Notice that $\vect{v}^\pi$ is an $m$-dimensional vector whose $i$-th component is the expected return of $\pi$ under the $i$-th objective.
\vspace{-1mm}
\paragraph{Solution concepts.} Unlike single-objective RL, where the value of a policy is a scalar value, comparing the values of two different policies in MORL is not a straightforward task. Indeed, in MORL, it is possible for a policy to excel in achieving higher expected return for one objective while simultaneously exhibiting lower performance for other objectives.
Hence, it is impossible to determine a total ordering on the value vectors, \textit{e.g.} how to know if $\vect{v}^{\pi} = (1,0)$ is better than $\vect{v}^{\pi'} = (0, 1)$ without knowing the user preferences with regards to the objectives at hand?

In such settings, algorithms usually rely on the concept of Pareto dominance as a partial ordering of the solutions. A policy $\pi$ is said to \textit{Pareto dominate} another policy $\pi'$ if its associated value vector is strictly greater for at least one objective without being worse for any other objective: 
\begin{equation*}
    \vect{v}^\pi \succ \vect{v}^{\pi'} \iff (\forall i : v^\pi_i \geq v^{\pi'}_{i}) \land  (\exists i : v^{\pi}_i > v^{\pi'}_{i}).
\end{equation*}

Since this ordering is partial, some solutions may not dominate or be dominated by others. A \textit{Pareto frontier} (PF) defines the set of all value vectors which are nondominated, representing all achievable trade offs between the objectives found by the MORL algorithm. Formally:
\begin{equation*}
     \mathcal{F} \equiv \{\vect{v}^\pi \ |\ \nexists \ \pi' \mathrm{ s.t. } \ \vect{v}^{\pi'} \succ \vect{v}^{\pi} \}.
\end{equation*}

This set is usually presented to the users after the learning process to let them choose which compromise to make before execution of the corresponding policy. Hence, \textbf{the solution returned by MORL algorithms} \textbf{consists in a set of policies} (that we denote as $\Pi^*$) linked to these nondominated value vectors, called \textit{Pareto set} (PS). 

In contrast, note that when the user preferences are known before the learning process (\textit{a priori}), MORL algorithms return only one policy since the user preferences provide a total ordering of the value vectors. Nevertheless, this study focuses on the setting where the user preferences are unknown. 

\vspace{-1mm}
\paragraph{Performance metrics} In the unknown preferences setting, it is  often challenging to evaluate the superiority of one algorithm over another because their respective PFs may (partially) overlap or intersect. In the field of multi-objective optimization and MORL, the use of \textit{performance metrics} is common practice. These metrics serve the purpose of transforming the PFs into scalar values, thereby facilitating comparison between algorithms. Two primary aspects are typically taken into consideration when assessing the quality of a PF: \textit{convergence} and \textit{diversity}. Convergence measures how closely the discovered PF aligns with an optimal PF, indicating the quality of the policies. Diversity, on the other hand, quantifies the variety of compromises presented, which allows for a broad range of trade-offs to be presented to the end user~\cite{talbi_metaheuristics_2009}. There exists many metrics that quantify either of those aspects, or even hybrid metrics quantifying both at the same time.

\textit{Inverted generational distance }(IGD)~\cite{coello_coello_study_2004} is a convergence metric that characterizes the distance between the PF found by the algorithm and a known optimal PF.  Let 
 $\mathcal{Z}$ be such a PF for the given domain. Then,
% \begin{equation*}
$    \mathrm{IGD}(\mathcal{F}, \mathcal{Z}) = \frac{1}{|\mathcal{Z}|}  \sqrt{\sum_{\vect{v}^* \in \mathcal{Z}}\min_{\vect{v}^\pi \in \mathcal{F}} \lVert \vect{v}^* - \vect{v}^\pi \rVert ^2}.$
% \end{equation*}

\textit{Sparsity}~\cite{xu_prediction-guided_2020} measures the diversity based on the distance between the policies found by the MORL algorithm. It can be computed using the following formula:
% \begin{equation*}
$\mathrm{S}(\mathcal{F}) = \frac{1}{|\mathcal{F}|-1} \sum_{j=1}^m\sum_{i=1}^{|\mathcal{F}|-1} (\mathcal{L}_j(i) - \mathcal{L}_j(i+1))^2,$
% \end{equation*}
where $\mathcal{L}_j$ is the sorted list of the values of the $j$-th objective considering all policies in $\mathcal{F}$, and $\mathcal{L}_j(i)$ is the $i$-th value in $\mathcal{L}_j$.

\textit{Hypervolume}~\cite{zitzler_evolutionary_1999} is a hybrid metric representing the volume of the area created between each point in the approximated Pareto front and a reference point in the objective space. This reference point, $\vect{v}_{\text{ref}}$, is (carefully) chosen to be a lower bound in each objective. Formally,
% \begin{equation*}
 $ \mathrm{HV}(\mathcal{F}, \vect{v}_{\text{ref}}) = \bigcup_{\vect{v}^\pi \in \mathcal{F}} \text{volume}(\vect{v}_{\text{ref}}, \vect{v}^\pi).$
% \end{equation*}

Finally, some metrics aim at quantifying the utility of the final user by modelling it using a function. In that vein, \textit{expected utility} (EU)~\cite{zintgraf_quality_2015} reflects the user utility across various potential preferences, under the assumption that it can be captured by a weighted sum utility function. Let $\Wspace$ be an $m$-dimensional simplex representing a set of weight vectors characterizing potential user preferences, then
% \begin{equation*}
$    \mathrm{EU}(\mathcal{F}) = \Ex_{\vect w\sim\Wspace} \left[\max_{\vect{v}^\pi\in\mathcal{F}} \vect{v}^{\pi} \cdot \vect w \right].$
% \end{equation*}

\vspace{-3mm}
\subsection{Hyperparameter optimization for RL}
\label{sec:HPO_RL}
% Background on HPO

Hyperparameter optimization for RL is the process of finding a set of hyperparameter values for the RL algorithm which leads to improved learning performances. The work of \cite{parker-holder_automated_2022,eimer_hyperparameters_2023} provide formal definitions of the objectives in hyperparameter optimization for single-objective RL. In general, for a given RL algorithm denoted $\mathrm A$, this problem can be defined as:

\begin{align} \label{eq:hporl}
\max_{\vect \zeta} \quad & f(\vect \zeta, \pi^{*}, v^{\pi^*}) \> \> \> \\ 
\text{s.t. } \quad & \pi^{*}, v^{\pi^*} = \mathrm{A}(\vect \zeta) \label{eq:rl} \\
& \vect \zeta \in Z \wedge \text{valid}(\vect \zeta), \label{eq:constraint}
\end{align}

where $\vect \zeta$ denotes the hyperparameters values, $Z$ denotes the space of hyperparameter values, $\pi^*$ and $v^{\pi^*}$ denote the (potentially optimal) policy found by the RL algorithm and its corresponding value\footnote{Notice that $v^\pi$ is just the counterpart of $\vect{v}^{\pi}$ in single-objective RL.}. $f$ denotes the objective function for the result of the learning process using the given hyperparameters.
%To emphasize on the difference between HPO and RL: the former focuses on finding the best hyperparameter values ($\vect \zeta^*$) for a given algorithm, whereas the latter learns the best policy ($\pi^*$) given a set of hyperparameter values (Equation~\ref{eq:hporl} and \ref{eq:rl}, respectively). 

Aside from finding hyperparameter values leading to good performance, it is interesting to remark that the data collected during the optimization process may be useful to understand the behaviour of the studied RL algorithm. Indeed, the search memory can be seen as a dataset where features consist of hyperparameters and labels are the resulting objective values. From this, one can study the relative importance of hyperparameters on the final performances using sensitivity analysis tools from supervised learning~\cite{bischl_hyperparameter_2023}.

While the mathematical definition of the problem seems straightforward, solving such an optimization problem is in practice very challenging due to the following aspects:

\vspace{-1mm}
\paragraph{Stochastic.} The effectiveness of RL algorithms can vary based on the random seeds employed during training~\cite{henderson_deep_2018,eimer_hyperparameters_2023}. As a result, aggregated metrics are often utilized to report these performances across multiple runs for statistical purposes. To achieve this in the HPO context, the RL algorithm needs to be executed on various seeds for each possible set of hyperparameter values and aggregated values of the resulting performance can be used as an objective function. Formally, 
\begin{equation}
\label{eq:stochastic}
f(\vect \zeta) = \mathrm{Agg}_{\sigma \in \Sigma_{\text{search}}}(\mathrm{Eval}(\mathrm{A}, \vect \zeta)),     
\end{equation}
where $\mathrm{Agg}_{\sigma \in \Sigma_{\text{search}}}$ represents an aggregation operator over a set of random seeds\footnote{To avoid confusion with the RL states, the seeds are denoted using sigma.} used in the search phase (typically the mean), and $\mathrm{Eval}$ gives the objective value of training one policy given the hyperparameter values. In general, this evaluation is linked to the value of the returned policy, \textit{e.g.} $\mathrm{Eval(A, \vect \zeta)} = v^{\pi^*}$, but more complex schemes exist and are discussed below.

Moreover, it is also crucial to ensure that the seeds utilized during the search phase are distinct from those employed for the final validation of the best hyperparameters discovered~\cite{eimer_hyperparameters_2023}, \textit{i.e.} $\Sigma_{\text{search}} \cap \Sigma_{\text{validation}} = \emptyset $. This precautionary measure prevents the optimization process from overfitting the search seeds.

\vspace{-1mm}
\paragraph{Multi-objective.} As mentioned above, the HPO objective function $f$ can be different depending on the context; it is in general linked to the agent performance on one environment, but it could also be the total training time required, or the performance of a policy on different environments if the user wants to optimize for policies able to generalize, \textit{e.g.} $\mathrm{Eval(A, \vect \zeta)} = \Ex_{i \sim \mathcal{I}} \left[ v^{\pi^*}(i) \right]$, with $\mathcal{I}$ representing a distribution of instances. Although most works choose to optimize only one objective for simplicity, some have already identified HPO for machine learning and RL to be multi-objective problems~\cite{parker-holder_automated_2022,jin_multi-objective_2006,binder_multi-objective_2020,horn_multi-objective_2016,bischl_hyperparameter_2023}.

\vspace{-1mm}
\paragraph{Constrained.} There is a potential for erratic behaviors and exceptions to occur during the RL learning process due to various combinations of hyperparameter values. In order to represent this aspect of the problem, Equation~\ref{eq:constraint} in the above optimization problem formulation establishes a space of configurations $Z$ that is then checked for validity. In practice, these configurations can be verified either within the optimizer or within the RL algorithm itself, for example by returning a penalized objective value if an invalid configuration is detected.

\vspace{-1mm}
\paragraph{Computationally expensive.} It is widely acknowledged that training an RL algorithm on a specific environment demands a significant amount of computing time. Additionally, it is important to note that during the HPO process, the RL algorithm needs to be executed with multiple random seeds for each potential combination of hyperparameter values. This practical requirement renders the problem impractical for exhaustive search methods.

In practice, such expensive problems are usually solved by relying on Bayesian Optimization (BO) \cite{mockus_bayesian_1975,jones_efficient_1998}, where a \textit{surrogate} model of the objective value is learned by the optimizer based on the collected samples during the search process. This guides the optimizer to choose the most promising areas of the hyperparameter space. Moreover, early-stopping mechanisms such as hyperband \cite{falkner_bohb_2018,li_hyperband_2018} help allocating more resources to well performing hyperparameters to better use the budget given to the optimization process.

\vspace{-1mm}
\section{Hyperparameter optimization for MORL}

The previous sections presented background information on the problems of multi-objective RL and hyperparameter optimization for RL. Building on these, this section presents the problem of hyperparameter optimization applied to the field of MORL.

\vspace{-1mm}
\subsection{Formal definition}
\label{sec:hpo_morl_def}

The main difference with HPO for RL lies in the fact that MORL algorithms return a Pareto set and its corresponding Pareto front. Hence, Equations~\ref{eq:hporl}--\ref{eq:constraint} can be adapted as follows:

\begin{align}
\max_{\vect \zeta} \quad & f(\vect \zeta, \Pi^{*}, \mathcal{F}) \> \> \> \\ 
\text{s.t. } \quad & \Pi^{*}, \mathcal{F} = \mathrm{A}(\vect \zeta) \label{eq:morl} \\
& \vect \zeta \> \in Z \wedge \text{valid}( \vect \zeta), 
\end{align}

where $\Pi^*$ and $\mathcal{F}$ represent the Pareto set of policies being learned by the MORL algorithm and its corresponding Pareto front. 

Dealing with sets as output entails an additional step in the HPO process: transforming the PFs into scalars for comparison. As discussed in Section~\ref{sec:MORL}, there are two aspects (or objectives) to consider in such a case: convergence and diversity. This means that the problem of HPO for MORL is also multi-objective. An initial simple solution to such an issue is to rely on performance metrics. In particular, hybrid performance metrics have the ability to quantify both aspects at the same time. Yet these metrics also suffer from their ability to summarize the information; for example, it is possible to have a large hypervolume score with only one well-placed point in the PF, while a PF composed by many points may have a small hypervolume score.

\vspace{-1mm}
\subsection{Proposed methodology}
% Figure of the train+validation process
% discuss each part

Building on all these aspects and extending the work of \cite{eimer_hyperparameters_2023}, we argue that the problem of HPO for MORL can be solved in two phases: a first phase which searches for hyperparameter values on a range of searching seeds, and a second phase which validates the final performance of the found hyperparameters on a range of validation seeds. 

Algorithm~\ref{algo:hpo} illustrates our proposed methodology. The algorithm starts by instantiating an optimizer, which is in charge of proposing hyperparameter values and keeping track of the associated final performances (line 1). Such an optimizer can be based on Bayesian Optimization such as \cite{falkner_bohb_2018}, but also could rely on simpler mechanisms such as grid search or random search. Next, the algorithm enters its search phase for a given amount of time specified by a stopping criterion. For each trial, the optimizer generates a new hyperparameter values candidate (line 3). These hyperparameters are then used to train the MORL algorithm on various search seeds for a given budget $b_{\text{search}}$ (lines 4--6). As outlined in Section~\ref{sec:hpo_morl_def}, a key distinction from HPO for RL emerges. In the MORL scenario, the outcomes of diverse runs manifest as PFs (sets of vectors), necessitating conversion into comparable entities. Consequently, each of these PFs are initially converted into scalar values through the utilization of a performance metric akin to those introduced in Section~\ref{sec:MORL}. Next, to deal with the stochastic aspect of the optimization problem, these result metrics are aggregated as discussed in Section 2.2 and reported to the optimizer (lines 7--8). Then, the hyperparameters leading to the best evaluation are returned by the optimizer to start the validation phase (line 11). In this phase, the MORL algorithm is trained over various validation seeds for a given budget $b_{\text{validation}}$ (lines 12--14). Finally, the validation performances and best-performing hyperparameters are returned to the user.

\begin{algorithm}
\caption{Hyperparameter optimization for MORL.\label{algo:hpo}}
\textbf{Input}: Stopping criterion $stop$, MORL algorithm $A$, hyperparameter space $Z$, environment $env$, searching budget $b_{\text{search}}$, validation budget $b_{\text{validation}}$, optimizer $Opt$, set of search seeds $\Sigma_{\text{search}}$, set of validation seeds $\Sigma_{\text{validation}}$.\\
\textbf{Output}: The best hyperparameter values for algorithm $A$ on $env$, $\vect \zeta^*$ and corresponding Pareto fronts.
\begin{algorithmic}[1]
\State $opt = \mathit{new\:Opt(}Z)$ \algorithmiccomment{Searching phase}
\While{$\neg \mathit{stop}$} 
    \State $ \vect \zeta = \mathit{opt.next()}$
    \For{$\sigma \in \Sigma_{\text{search}}$}
        \State $\mathcal{F}_\sigma, \Pi^*_\sigma = A(\vect \zeta, \sigma, b_{\text{search}}, env)$
    \EndFor
    \State $f = \mathit{Agg_{\sigma \in \Sigma_{\text{search}}}(Eval(}\mathcal{F_\sigma}))$ \algorithmiccomment{See Equation~\ref{eq:stochastic}}
    \State $\mathit{opt.report(}\vect \zeta, f)$
\EndWhile \\

\State $\vect \zeta^* = \mathit{opt.best()}$ \algorithmiccomment{Validation phase}
\For{$\sigma \in \Sigma_{\text{validation}}$} 
    \State $\mathcal{F}_\sigma, \Pi^*_\sigma = A(\vect \zeta^*, \sigma, b_{\text{validation}}, env)$
\EndFor
\State \textbf{return }$\mathcal{F}, \vect \zeta^*$
\end{algorithmic}
\end{algorithm}

% \begin{figure}
%     \centering
%     \includegraphics[width=0.4\textwidth]{images/hpoMORL.png}
%     \caption{Diagram of the proposed methodology.}
%     \label{fig:hpo_morl}
% \end{figure}
\vspace{-1mm}
\section{Experiments}
% say that we are currently building the tool but do not have the full picture
% Justify why wandb: already integrated, good ui, tools for posterior analysis
% Describe minecart problem: it requires a lot of exploration to reach the mines and come back to sell the ores: sparse rewards w.r.t. ores and delayed
% Parameter importance and correlation tool
This section presents some early results of the proposed methodology on a state-of-the-art MORL algorithm. Our main goals are: (1) to validate that HPO provides an advantage compared to relying on default hyperparameter values, (2) to showcase the potential benefits of studying hyperparameters through the lense of sensitivity analysis on the search memory.

\begin{table}[b]
\renewcommand{\arraystretch}{1.2}
\begin{center}
{\caption{Hyperparameters search space ($Z$).}\label{tab:hyperparams_space}}
\begin{tabular}{lc}
\hline
Hyperparameter & Value range
\\
\hline
\texttt{learning\_rate} & $[0.001, 0.0001]$ \\
\texttt{learning\_starts} & $[1..1,000]$ \\
\texttt{buffer\_size} & $[1,000..2,000,000]$ \\
\texttt{max\_grad\_norm} & $[0.1, 10.0]$ \\
\texttt{gradient\_updates} & $[1..10]$ \\
\texttt{target\_net\_update\_freq} & $[1..10,000]$ \\
\texttt{tau} & $[0.0, 1.0]$ \\
\texttt{num\_sample\_w} & $[2..10]$ \\
\texttt{per\_alpha} & $[0.1, 0.9]$ \\
\texttt{initial\_homotopy\_lambda} & $[0.0, 1.0]$ \\
\texttt{final\_homotopy\_lambda} & $[0.0, 1.0]$ \\
\texttt{homotopy\_decay\_steps} & $[1..100,000]$ \\
\texttt{initial\_epsilon} & $[0.01, 1.0]$ \\
\texttt{final\_epsilon} & $[0.01, 1.0]$ \\
\texttt{epsilon\_decay\_steps} & $[1..100,000]$ \\

\hline
\end{tabular}
\end{center}
\end{table}

In practice, we optimized 15 hyperparameters of Envelope Q-Learning \cite{yang_generalized_2019} according to the hyperparameter space described in Table~\ref{tab:hyperparams_space}. Having such a large number of hyperparameters to consider makes it virtually impossible for a human to tackle the problem. We study this algorithm on the \textit{minecart-v0} environment~\cite{abels_dynamic_2019,alegre_mo-gym_2022}. In this environment, the agent controls a cart and has three objectives: it must collect two types of ore from different mines as well as minimize fuel consumption. Finding a diverse PF in such a domain is particularly challenging due to the delayed rewards associated with both ores. Indeed, the agent must first go to a mine to gather ore, then return to its base to obtain the related rewards. This necessitates a significant amount of exploration.

\begin{table}[t]
\renewcommand{\arraystretch}{1.2}
\begin{center}
{\caption{HPO parameters for our experiments.}\label{tab:params}}
\begin{tabular}{lc}
\hline
Parameter & Value
\\
\hline
$Z$ & See Table~\ref{tab:hyperparams_space}\\
$\mathit{stop}$ & 48 hours (37 trials)\\
$\Sigma_{\text{search}}$ & $[10..12]$ \\
$b_{\text{search}}$ & 100,000 steps \\
$\mathit{Eval}$ & $\mathrm{HV}(\mathcal{F}_\sigma$), $\vect{v}_{\text{ref}}$ = (-1,-1,-200) \\
$\mathit{Agg}$ & Mean \\
$\Sigma_{\text{validation}}$ & $[0..9]$ \\
$b_{\text{validation}}$ & 400,000 steps \\

\hline
\end{tabular}
\end{center}
\end{table}

\begin{figure*}[t]
    \centering
    \includegraphics[width=0.93\textwidth]{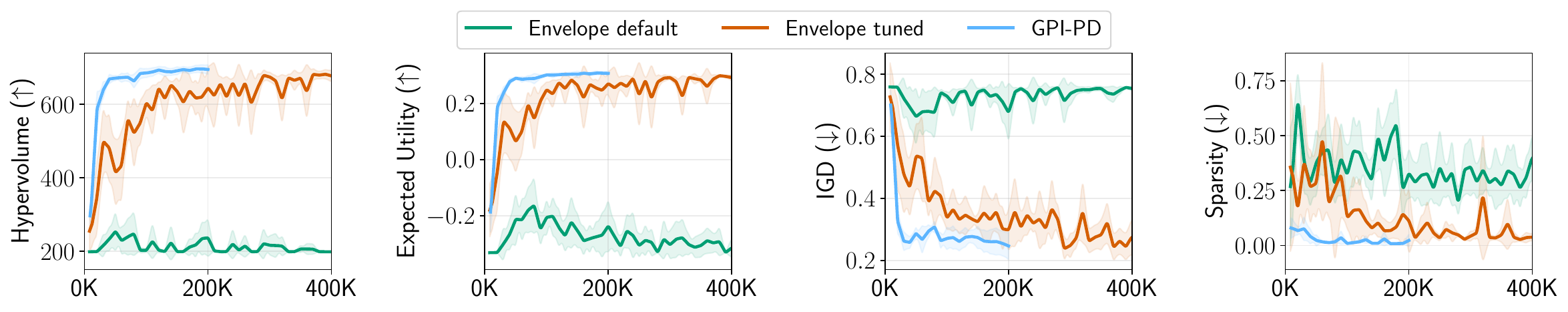}
    \caption{Average performance metrics and 95\% confidence intervals over 10 seeded runs of the tuned algorithm and existing baselines on \textit{minecart-v0}.}
    \label{fig:validation_results}
\end{figure*}

Our experiments were conducted using the implementations in MORL-Baselines~\cite{felten_toolkit_2023}, MO-Gymnasium~\cite{alegre_mo-gym_2022}, and we used the Bayesian optimizer of Weights and Biases (W\&B) \cite{biewald_experiment_2020}. Table~\ref{tab:params} presents a global overview of the parameters linked to the optimization process\footnote{The W\&B hyperparameter optimization dashboard can be found at: \url{https://api.wandb.ai/links/openrlbenchmark/9ffv3e5c}. Our code can be found at \url{https://github.com/LucasAlegre/morl-baselines/tree/paper/hpo-morl-modem}.}. To tackle the stochastic aspect of the problem, we chose to run each trial on 3 search seeds (in parallel) in order to balance between speed and accuracy of the trials' estimates. Hypervolume has been chosen as performance metric because of its ability to characterize both diversity and convergence. The total optimization budget was 2 days, which was equivalent to running $37 \times 3$ trainings (each one for 100,000 steps). For reference, each validation training (400,000 steps) required approximately 3.5 hours.
All our experiments have been conducted using the high-performance computer of the University of Luxembourg~\cite{varrette_management_2014}. 

% Figure~\ref{fig:wandb} presents the resulting dashboard of the optimization process (first phase). The interface allows to find the hyperparameters leading to the highest average hypervolume using a parallel coordinate plot, but also to study the importance and correlation of the hyperparameters w.r.t the objective. In particular, it is interesting to notice that \texttt{epsilon\_decay\_steps} reveals to be the most important hyperparameter. This parameter designs the final exploration rate of the agent

\vspace{-1mm}
\subsection{Sensitivity analysis}

The integration with W\&B provides dashboards allowing  analysis of the search phase. These dashboards utilize the search memory as a dataset, where hyperparameter values serve as features and resulting performances as labels. Then, a random forest is trained to calculate parameter importance, while linear correlations between hyperparameters and final results are also presented. In the case under study, we extracted the four most significant hyperparameters from the dashboard and listed these in Table~\ref{tab:analysis}. 

\begin{table}
\renewcommand{\arraystretch}{1.2}
\begin{center}
{\caption{Hyperparameters importance and correlations.}\label{tab:analysis}}
\begin{tabular}{lcc}
\hline
Parameter & Importance & Correlation
\\
\hline
\texttt{final\_homotopy\_lambda} & $0.288$ & 0.678 \\
\texttt{tau} & $0.222$ & -0.617 \\
\texttt{homotopy\_steps} & $0.101$ & 0.558 \\
\texttt{epsilon\_decay\_steps} & $0.064$ & 0.457 \\
\hline
\end{tabular}
\end{center}
\end{table}

The table indicates that the homotopy optimization approach used by Envelope has a strong influence on the final results. This involves optimizing a linear combination of two loss functions to smoothen the objective landscape, given by $L(\theta) = (1-\lambda_k) \cdot L^{\mathtt A}(\theta) + \lambda_k \cdot L^{\mathtt B}(\theta)$, where $L^{\mathtt A}$ is a loss based on vectorial Q-values and $L^{\mathtt B}$ is based on scalarized values. The parameter $\lambda_k \in [$\texttt{initial\_homotopy\_lambda}, \texttt{final\_homotopy\_lambda}$]$ is increased throughout the learning process based on a schedule given by the parameters listed in the table. It seems that on this environment, a large $\lambda$ (and thus relying more on $L^{\mathtt B}$) is correlated with good performances.

The parameter \texttt{tau} is linked to the soft update of the target network, as in DQN~\cite{mnih_human-level_2015}. Keeping this parameter relatively low appears to be linked to better performance. Finally, \texttt{epsilon\_decay\_steps} regulates the decay rate of the agent's exploration strategy. Maintaining a high level of exploration appears to be associated with favorable final performance. This observation is consistent with the earlier discussion concerning the environment's delayed rewards. For a more detailed explanation of these parameters, refer to the original Envelope paper by Yang et al.~\cite{yang_generalized_2019}.

% \begin{figure}
%     \centering
%     \includegraphics[width=0.45\textwidth]{images/wandb.png}
%     \caption{Dashboard of hyperparameter search results, with panel for analysis of parameter importance and correlation.}
%     \label{fig:wandb}
% \end{figure}

\vspace{-1mm}
\subsection{Validation}

% \begin{table}
% \renewcommand{\arraystretch}{1.2}
% \begin{center}
% {\caption{Best hyperparameter values found.}\label{tab:best_hyperparams}}
% \begin{tabular}{lc}
% \hline
% Hyperparameter & Value range
% \\
% \hline
% \texttt{learning\_rate} & $0.00024$ \\
% \texttt{learning\_starts} & $167$ \\
% \texttt{buffer\_size} & $1,486,469$ \\
% \texttt{max\_grad\_norm} & $1.2262$ \\
% \texttt{gradient\_updates} & $5$ \\
% \texttt{target\_net\_update\_freq} & $3,022$ \\
% \texttt{tau} & $0.1294$ \\
% \texttt{num\_sample\_w} & $3$ \\
% \texttt{per\_alpha} & $0.3036$ \\
% \texttt{initial\_homotopy\_lambda} & $0.9021$ \\
% \texttt{final\_homotopy\_lambda} & $0.8728$ \\
% \texttt{homotopy\_decay\_steps} & $51,843$ \\
% \texttt{initial\_epsilon} & $0.7293$ \\
% \texttt{final\_epsilon} & $0.5540$ \\
% \texttt{epsilon\_decay\_steps} & $95,463$ \\

% \hline
% \end{tabular}
% \end{center}
% \end{table}

To validate the effectiveness of the proposed method, we compare the results of training Envelope with the tuned hyperparameters (Envelope tuned) against training with the default hyperparameters (Envelope default) defined in MORL-Baselines. Furthermore, to establish an upper baseline, we compare the results against GPI-PD~~\cite{alegre_sample-efficient_2023}, one of the most sample-efficient MORL algorithm available at the time of writing. For the baseline results, we reuse the ones provided by the MORL-Baselines authors in openrlbenchmark~\cite{huang_openrlbenchmark_2023}. To compare these algorithms, we rely on the performance metrics discussed in Section~\ref{sec:MORL}. These allow assessing the diversity and convergence aspects of the resulting PFs as well as the final expected user utility.

Figure~\ref{fig:validation_results} illustrates the learning curves of the three different algorithms over training steps. There is a drastic difference between Envelope default and its tuned counterpart. Indeed, Envelope default seems to fail to discover new policies over the course of the learning process, as all metrics stay rather flat. On the other hand, Envelope tuned improves on all metrics, meaning it improves the learned PF. Upon comparison with GPI-PD, Envelope tuned exhibits remarkably strong performance, achieving results almost on par with this more recent technique. It is worth noting that Envelope benefits from a faster implementation in practice, resulting in even closer performance when measured against walltime instead of environment steps. Naturally, to ensure a fair comparison between Envelope and GPI-PD, the latter algorithm should also be tuned on this environment. 
\vspace{-1mm}
\section{Conclusion}

In this work, we discussed the benefits and challenges of hyperparameter optimization for multi-objective reinforcement learning. We presented the existing literature on MORL and HPO for RL. Then, we derived a methodology for conducting HPO for MORL based on these. As early results, our methodology has been applied on a state-of-the-art algorithm and shown to greatly improve its final performances on a classical MORL environment. We also present an analysis of the importance of some hyperparameters using the newly introduced tools. This work is opening avenues for a significant amount of future research opportunities.

First, our implementation could be further improved in various ways. For example, one could improve the optimizer by including more advanced techniques such as hyperband~\cite{li_hyperband_2018} or by distributing the search phase over various computing nodes. Moreover, the set of tools for sensitivity analysis of the optimization process to gain better understanding of the hyperparameters' influence could be enhanced by relying on the extensive literature from the supervised learning community. 

Second, it is important to evaluate the adequacy of utilizing three search seeds as a general practice. This assessment involves confirming whether the optimization consistently identifies comparable values for high-performing hyperparameters across multiple runs employing varying search seeds.

Third, an avenue for exploration could involve an investigation into the generality of the discovered hyperparameters. This exploration would entail determining whether the identified hyperparameters yield favorable performance exclusively within the studied environment, across a range of related environments (e.g., Mujoco~\cite{todorov_mujoco_2012}), or even universally across all environments.

Fourth, this work could also be extended to automated algorithm design, where entire algorithmic choices, such as the Bellman update components, can be decided by the optimizer.

Finally and more importantly, it would be interesting to apply the proposed methodology to a broader set of applications, \textit{e.g.} other algorithms and other environments, and study its results. In particular, we believe that the process of HPO for MORL will play a pivotal role in methodically accumulating significant empirical evidence regarding MORL algorithms' performance. This, in turn, will aid in enabling a thorough comparative analysis of various MORL algorithms.

\vspace{-2mm}
\section*{Acknowledgements}
This work was financed by the Fonds National de la Recherche Luxembourg (FNR), CORE program under the ADARS Project, ref. C20/IS/14762457.

\clearpage
\bibliography{bib/main}
\end{document}